\documentclass[journal,twoside,web]{ieeecolor}

\usepackage{jsen}
\usepackage{cite}
\usepackage{amsmath,amssymb,amsfonts}
\usepackage{algorithm}
\usepackage{algpseudocode}
\usepackage{graphicx}
\usepackage{textcomp}
\usepackage{wrapfig}
\usepackage{siunitx}
\usepackage{romannum}

\makeatletter
\renewcommand{\ALG@beginalgorithmic}{\small}
\makeatother

\def\BibTeX{{\rm B\kern-.05em{\sc i\kern-.025em b}\kern-.08em
    T\kern-.1667em\lower.7ex\hbox{E}\kern-.125emX}}
\definecolor{abstractbg}{rgb}{0.89804,0.94510,0.83137}
\begin{document}
\title{A Neuromorphic Electronic Nose Design}
\author{Shavika Rastogi, Nik Dennler, Michael Schmuker, and André van Schaik, \IEEEmembership{Fellow, IEEE}
\thanks{Part of this work was funded by an NSF/MRC award under the Next Generation Networks for Neuroscience initiative (NeuroNex Odor to action, NSF \#2014217, MRC \#MR/T046759/1)}
\thanks{Shavika Rastogi and Nik Dennler are with the International Centre for Neuromorphic Systems, Western Sydney University, Kingswood 2747 NSW, Australia, and also with Biocomputation Group, University of Hertfordshire, Hatfield AL10 9AB, United Kingdom (e-mail: rastogi.shavika@gmail.com; dennler@proton.me)}
\thanks{Michael Schmuker is with Biocomputation Group, University of Hertfordshire, Hatfield AL10 9AB, United Kingdom and also with BioMachineLearning Research Consulting, Berlin, Germany (e-mail: m.schmuker@biomachinelearning.net)}
\thanks{André van Schaik is with the International Centre for Neuromorphic Systems, Western Sydney University, Kingswood 2747 NSW, Australia (e-mail: a.vanschaik@westernsydney.edu.au).}}

\IEEEtitleabstractindextext{%
\fcolorbox{abstractbg}{abstractbg}{%
\begin{minipage}{\textwidth}%
\begin{wrapfigure}[12]{r}{2.7in}%
\includegraphics[width=2.5in]{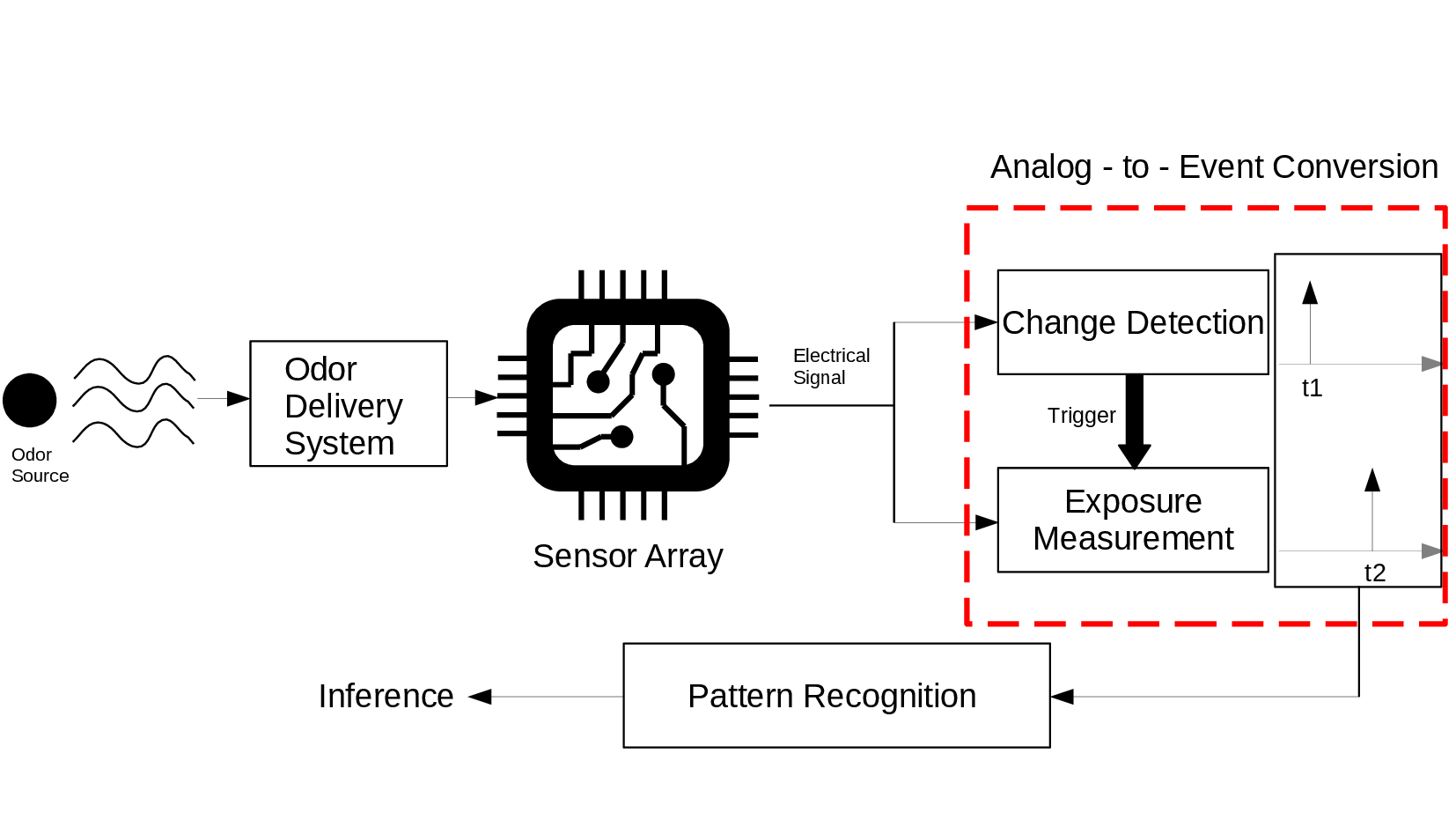}%
\end{wrapfigure}%
\begin{abstract}
Rapid detection of gas concentration is important in different domains like gas leakage monitoring, pollution control, and so on, for the prevention of health hazards. Out of different types of gas sensors, Metal oxide (MOx) sensors are extensively used in such applications because of their portability, low cost, and high sensitivity for specific gases. However, how to effectively sample the MOx data for the real-time detection of gas and its concentration level remains an open question. Here we introduce a simple analog front-end for one MOx sensor that encodes the gas concentration in the time difference between pulses of two separate pathways. This front-end design is inspired by the spiking output of a mammalian olfactory bulb. We show that for a gas pulse injected in a constant airflow, the time difference between pulses decreases with increasing gas concentration, similar to the spike time difference between the two principal output neurons in the olfactory bulb. The circuit design is further extended to a MOx sensor array and this sensor array front-end was tested in the same environment for gas identification and concentration estimation. Encoding of gas stimulus features in analog spikes at the sensor level itself may result in data and power-efficient real-time gas sensing systems in the future that can ultimately be used in uncontrolled and turbulent environments for longer periods without data explosion. 
\end{abstract}

\begin{IEEEkeywords}
Metal oxide sensors, olfactory bulb, analog front-end, gas concentration. 
\end{IEEEkeywords}
\end{minipage}}}

\maketitle

\section{Introduction}
\label{sec:introduction}
\IEEEPARstart{A}{ll} creatures perceive this world and interact with the environment through different types of senses, where the nervous system extracts some meaningful features from the external stimulus and uses these to build an internal representation of the environment. The sensory information present in the environment is sampled and encoded by the nervous system in such a way that it can be easily processed, learned, and efficiently transmitted to the appropriate region of the brain so that the animal can respond to its environment quickly enough, for example, to avoid danger or to catch prey. The understanding of sensory processing in our nervous system can help us make existing man-made systems more data efficient and operate in real-time. \par

Gas sensors are widely used for the detection of gas and its concentration in various domains: industrial production (for example, methane detection in mines), automotive industries (for example, detection of polluting gases from vehicles), medical applications, indoor air quality monitoring (for example, gas leakage detection) and environmental monitoring \cite{Liu2012}. Metal oxide (MOx) gas sensors are widely used in portable gas detection systems because of their low cost, quick response time, easy production, simple peripheral electronics, high sensitivity, and high selectivity to many gases \cite{Liu2012}\cite{Nikolic2020}. A conventional electronic nose (e-nose) \cite{Vanarse2017} contains an odor delivery system and an array of MOx sensors forming an analog sensing unit. The analog sensing unit interacts with the environment to collect odor molecules and convert them into electrical signals. These electrical signals are then sent to the processing unit of the e-nose for further processing, which includes analog-to-digital converters (ADCs), sensor data preprocessing units, and a memory-based pattern recognition engine \cite{Vanarse2017}\cite{Gardner1994}. \par

According to the components of an e-nose system, the performance of an e-nose can be improved by optimizing: (1) the selection of the sensitive materials and sensor array optimization, (2) the feature extraction and selection method, and (3) the pattern recognition method \cite{Yan2015}. For a commercial e-nose \cite{Lebrun2007}\cite{Hudon2000}, we cannot change the sensor array, but we can optimize the feature extraction and pattern recognition method \cite{Yan2015}. The pattern recognition unit contains many computational stages where significant features of sensor response are extracted and utilized for odor identification. Cheng et al. \cite{Cheng2021} have listed different pattern recognition methods that have been used to process the sensor data. \par

For most pattern recognition methods, the sensor data has to be sampled to be used by a computer or microprocessor. ADCs typically employed in an e-nose system usually have a fixed sampling frequency. Therefore, the samples are generated at a fixed time interval throughout the operation of the e-nose system, irrespective of the presence or absence of any gas in the environment. This generates a large number of irrelevant samples to deal with at the pattern recognition stage which increases the computational cost. In biological neural systems, data is transmitted between two nodes only when it carries relevant information. %If we look at biological neural systems, the data is transmitted from one point to another only when there is some relevant information. 
This leads to significant data reduction and makes the inference easier. \par

Inspired by the efficiency of biological information processing systems and the effectiveness of neurobiological solutions despite ill-conditioned sensory input data, Carver Mead \cite{Mead1990} proposed the idea of Neuromorphic Engineering. The primary objective of Neuromorphic Engineering is to devise technologies that draw inspiration from the brain and enable the extraction of meaningful information from the external world \cite{Mead1990}\cite{Lyon88}. In the emerging field of Neuromorphic Olfaction, the computational principles of biological olfactory systems are studied and translated into algorithms and devices \cite{Chicca2014}. Over the years, the field of neuromorphic olfaction has grown in different directions. Neuromorphic methods have been developed in the field of artificial olfaction which tried to emulate the biological olfactory pathway \cite{Beyeler2010, Imam2012, Rivera2007, Raman2004}, or to investigate some meaningful signal processing principles for gas sensor data \cite{dennler2022}. Some works also tried to emulate only those neurobiological principles that can be implemented in silicon \cite{Pan2012, Hung2012, Hausler2011}. Other works were focused on improving the performance of gas-sensing systems. Therefore, this field also observed several modular developments in which the primary objective was to either develop a sensory front-end with spike-based output \cite{Covington2003, Marco2014, Huang2017} or a spike-based processing unit \cite{Tang2011, Kasap2013, Martinelli2009, Hoda2010}. Recently, neuromorphic olfactory sensors have also been developed for some specific applications \cite{Jang2024}\cite{Song2024}. \par

Spike-based gas sensing front-ends extract and encode gas features (e.g.: gas identity) from the sensor data in the form of spikes, simplifying further processing. Gas sensing front-ends that have been implemented so far convert the output of MOx sensors into concentration-invariant spike patterns \cite{Huang2017}\cite{Ng2011}\cite{Yamani2012}.  However, the determination of gas concentration apart from gas identity is also important in many applications such as for the development of robotic plume navigation systems \cite{Chen2019}, where the temporal features of odor plumes \cite{Conchou2019} might be useful for odor source localization, or for indoor air monitoring \cite{MUMYAKMAZ2015} where severe exposure to a toxic gas might be dangerous. In such applications, there exists a demand for rapid gas concentration detection systems so that appropriate action can be taken on time. Han et al. \cite{Han2022} proposed an artificial olfactory neuron module comprising a chemo-resistive gas sensor and a one-transistor neuron. The gas concentration was encoded in the spiking frequency of the olfactory neuron module. This implementation was an important step towards encoding additional information about gas in spikes apart from gas identity using neuromorphic methods. \par

The primary goal of the work presented in this article is to build a simple sensory front-end using mixed-signal CMOS circuits for existing gas sensing systems that can encode the gas features such as its identity and concentration level in the environment into events similar to the olfactory bulb of mammals. The olfactory bulb is a pre-processing station that is known to extract interesting features out of the raw odor signal before the olfactory cortex comes into play for the recognition process \cite{Mori1999}\cite{Kay2006}. Modeling the mechanisms inside the olfactory bulb using CMOS circuits for gas sensing systems might help us to develop an encoding scheme to encode the gas sensor data into events such that useful features (e.g.: identity and concentration) about the gas are preserved \cite{Rastogi2023}. The response time of MEMS-based MOx gas sensors available today lies in the range of seconds \cite{Asri2021}. Recently, Dennler et al. \cite{dennler2024} have shown that a high-speed e-nose comprising MEMS-based MOx sensors with rapid heater modulation can resolve odor pulses in the range of milliseconds which is at par with the temporal resolution of animal olfaction. To account for the time constants in the range of milliseconds to seconds, the capacitances needed for a gas-sensing circuit will be large and are usually bulky for a semiconductor chip. Therefore, a gas-sensing circuit is more suitable for a PCB-level design rather than a chip-level design. We first designed the circuit for a single sensor in a controlled environment with a single gas pulse. The proposed circuit for a single sensor can encode the concentration level of a known gas in spike timings. We then extended the circuit design for a 3 sensor array in the same environment. The gas identity and concentration level can be estimated using the combinatorial output of sensors in the sensor array circuit.  

\section{Materials and Methods}

\subsection{Experimental Setup for E-nose Recordings}
The data collection campaign using a
custom-made e-nose % \cite{dennler2024} was used for data collection and it comprises f
has been described in a previous study \cite{dennler2024}, and leverages 
four different MOx gas sensors operated at a constant heater voltage. Out of these sensors, reducing (RED) (Sensor 1), oxidizing (OX) (Sensor 2), and Ammonia ($NH_{3}$) (Sensor 3) sensors of MiCS-6814 have been used for our circuit design. For the data collection, the odor stimuli were provided by a multi-channel odor delivery device described in detail by Ackels et al. \cite{Ackels2021}. It offers an exceptionally high temporal fidelity, which is achieved by combining high-speed gas valves, flow controllers, as well as short and narrow gas pathways. Odorant headspace samples were embedded in a constant airflow and presented to the e-nose. The delivered odor concentration was varied by modulating the valve shatter duty cycle, ensuring a linear relationship between different concentration levels. Fluctuations in the flow during gas presentations were minimized by careful calibration. For our design, recordings of three different odor stimuli have been used: Ethyl Butyrate (EB), Eucalyptol (Eu), and Isoamyl Acetate (IA). These gases were diluted in an odorless mineral oil at a ratio of 1:5. \par

\begin{figure}[h]
\includegraphics[width=\linewidth]{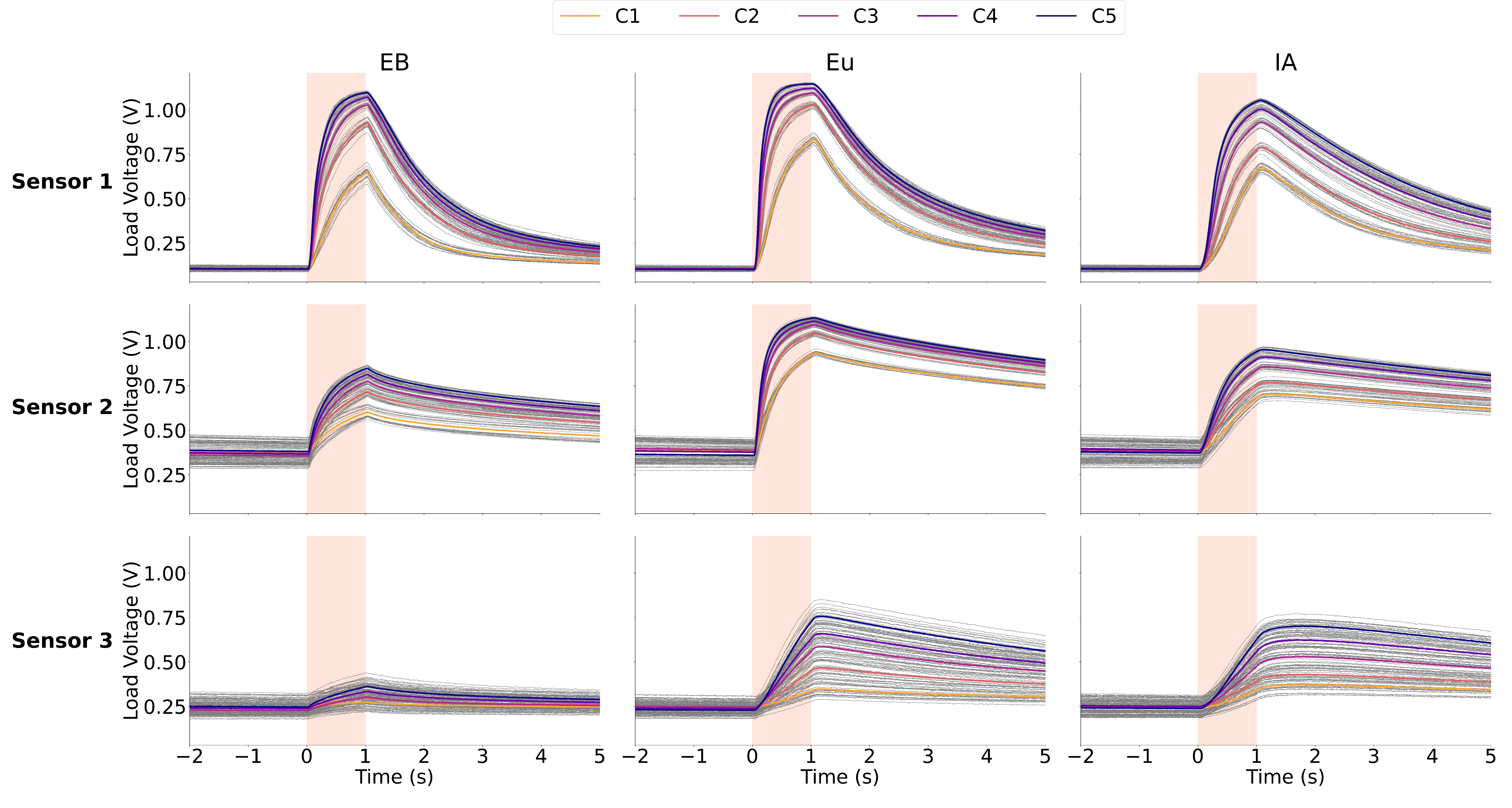}
\caption{Mean load voltage over all trials obtained for three different gases at five different concentration levels for three sensors of MiCS-6814. Gas stimulus is indicated by the shaded region, with the gas type indicated in the title of each panel. C1 to C5 indicate 5 concentration levels of each gas such that C1 is the lowest and C5 is the highest concentration level, and each colored curve indicates the mean response at these concentrations. The individual trials are shown in grey.}
\label{fig:1_conc_graph}
\end{figure}

Figure 1 shows the mean sensor measurements over 20 trials for three odor stimuli at five different concentration levels for three sensors of MiCS-6814. The relative gas concentration at these concentration levels was 20 \% (C1), 40 \% (C2), 60 \% (C3), 80 \% (C4), and 100 \% (C5). The measurement consists of voltage recordings across the load resistors connected in series with sensors. This voltage varied inversely with respect to MOx sensor resistance and was used as the input signal to the circuit. The load resistance values connected to sensors 1, 2 and 3 are $27 k\Omega$, $6.8 k\Omega$, and $27 k\Omega$ respectively. Negative timestamps indicate the sensor's baseline response before gas release. \par

At $t=0 s$ the gas is released and continued for 1s (as indicated by the shaded region). Gas release stopped at $t=1 s$ and sensors were allowed to return to baseline for 30s before the subsequent consecutive trial of the experiment started. Each experiment was repeated 20 times, where the order of odors and their concentrations were randomized as described in \cite{dennler2024}.

\subsection{Dual Pathways inside Mammalian Olfactory Bulb}

Mitral cells (MC) and Tufted cells (TC) are the two principal output neurons inside the olfactory bulb which are responsible for the transmission of the pre-processed odor information in the form of spike trains from the olfactory bulb to the olfactory cortex. These two neuron types can be clearly distinguished by their dendritic morphologies \cite{Haberly1977} and also by their locations inside the olfactory bulb \cite{Mori1983}\cite{Orona1984}. Also, these neurons project to two distinct but overlapping parts of the olfactory cortex. The axonal projections of MCs cover the entire piriform cortex while those of TCs are restricted to the anterior part of the piriform cortex and more rostral structures \cite{Haberly1977}\cite{Nagayama2010}. These findings indicate that even though MCs and TCs receive similar odor signals from a shared glomerulus, the odor information is processed differently by these two neurons. \par

It has been reported that MCs and TCs fire in two distinct and opposite phases of the sniff cycle, as shown in Figure 2 (a). The depolarization of MCs occurs during the inhalation period and for TCs, it happens during the exhalation period, which leads to the distinct preferred phases of action potential discharge for these two neuron types \cite{Fukunaga2012}. It has been found that with the increase in odor concentration (as depicted in Figure 2 (b)), the firing rate of TCs increases without any phase change. But MCs start firing earlier in the sniff cycle in addition to an increased firing rate. In this way, the odor concentration is encoded in the phase (or time) difference between the spiking activities of MCs and TCs. \par

\begin{figure}[h]
\includegraphics[width=\linewidth]{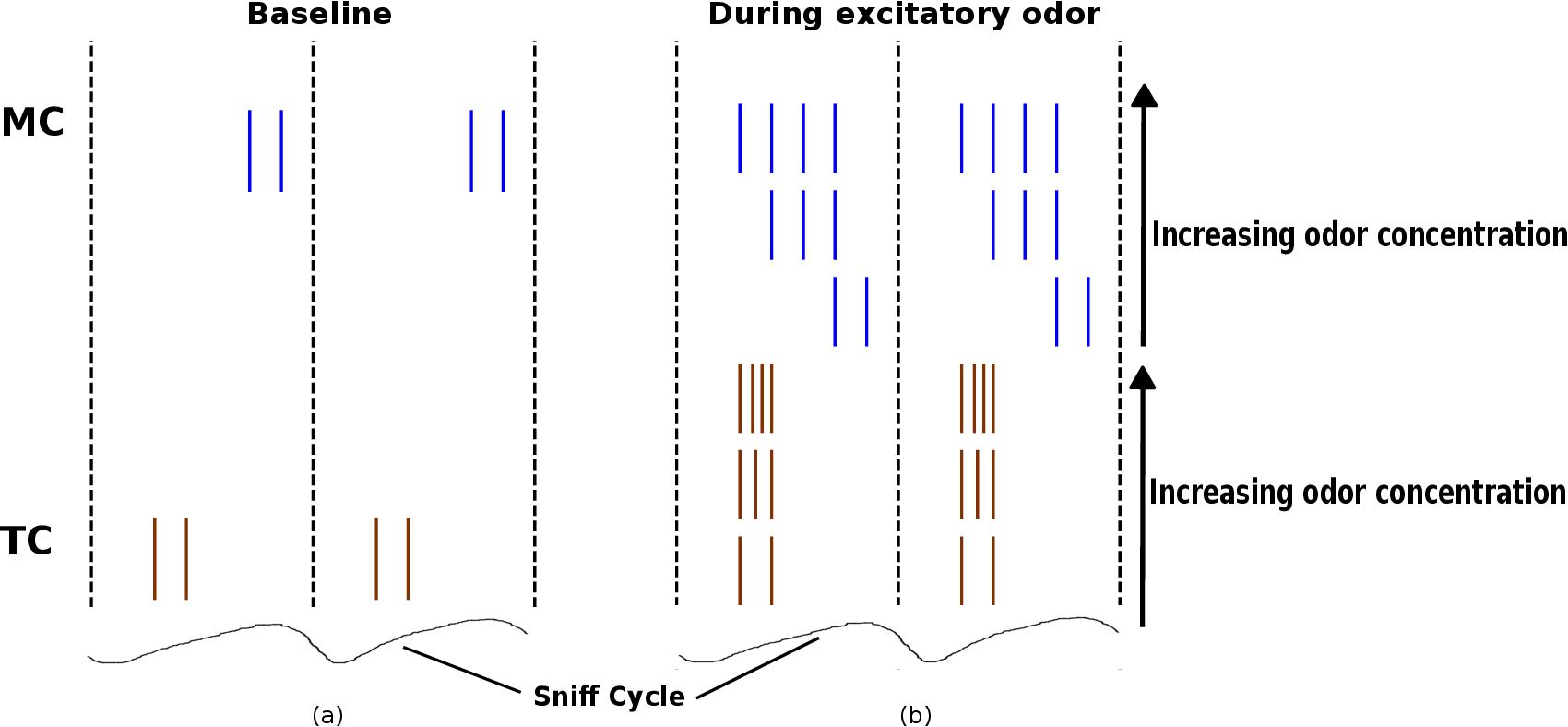}
\caption{(a) Baseline firing of MCs and TCs. Both cells fire in distinct and opposite phases of the sniff cycle. (b) Variation in MC and TC firing with increasing odor concentration, as indicated by arrows. Figure adapted from \cite{Fukunaga2012}}
\label{fig:mc_tc_profile}
\end{figure}

\section{Proposed Circuit Design}

\subsection{Single Sensor}
 Taking inspiration from parallel pathways inside the mammalian olfactory bulb, we have designed an analog circuit for a single MOx sensor (Sensor 1) for encoding gas concentration levels in the timings of analog spikes emerging out of two parallel pathways. We have used the Skywater 130 Process Design Kit (PDK) \cite{edwards2021introduction} for the design and simulation of all circuits. As it is a chip-designing tool, it allows us to build different components, like op-amp and inverters, from individual transistors. This tool also allows us to adjust the sizing of individual transistors to get the desirable response from the circuit. However, as the practical number of MOx channels available in an e-nose is small, they do not require a chip-based readout, and our design is aimed at implementation using discrete components on a Printed Circuit Board. \par

\begin{figure}[h]
\includegraphics[width=\linewidth]{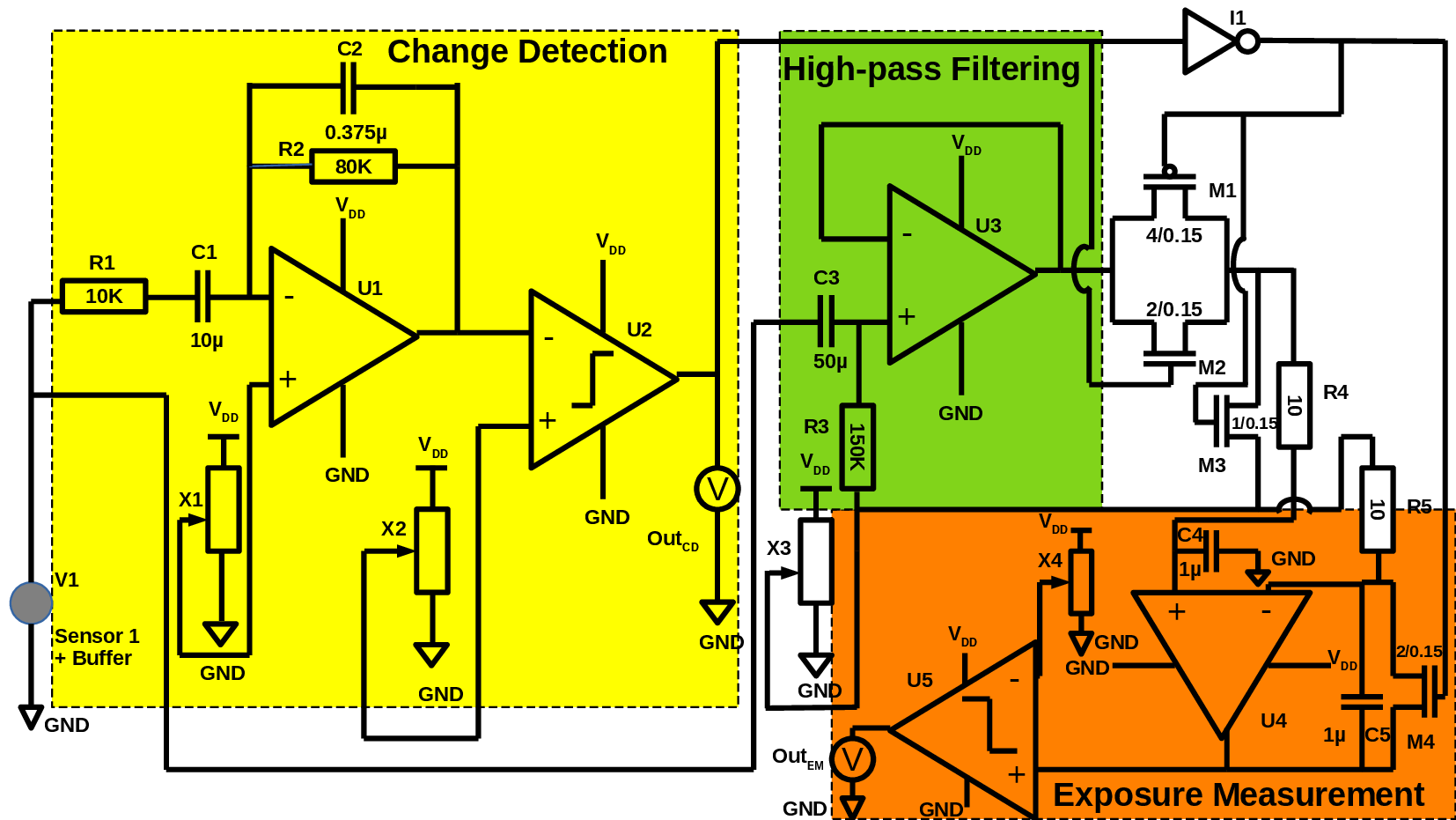}
\caption{The proposed gas concentration measurement circuit for a single MOx sensor.}
\label{fig:circuit_front_end_single_sensor}
\end{figure}

Figure 3 shows the proposed analog front-end circuit design for the single MOx sensor (Sensor 1). It consists of two sub-circuits: (1) A Change Detection circuit, and (2) An Exposure Measurement circuit.

\subsubsection{The Change Detection Circuit}
The change detection circuit consists of an inverting op-amp difference circuit U1 with a band-pass filtering operation followed by an op-amp comparator U2. The MOx sensor is connected in series with a load resistor. The voltage drop V1 across this load serves as an input to the circuit. The inverting op-amp difference circuit U1 is provided with the DC offset at its non-inverting terminal which is adjustable by the potentiometer X1. The gain of the circuit depends on resistor R2 and capacitor C1. All resistors and capacitors used in the circuit together determine the rise and decay time of the circuit output. \par

The inverting difference circuit output is compared with a pre-defined threshold voltage by the comparator U2. A change detection pulse ($Out_{CD}$) is generated whenever the output of the differentiator exceeds this threshold. This threshold voltage level is adjustable by the potentiometer X2 and can be varied depending on the SNR of the input signal.

\begin{figure}[h]
\includegraphics[width=\linewidth]{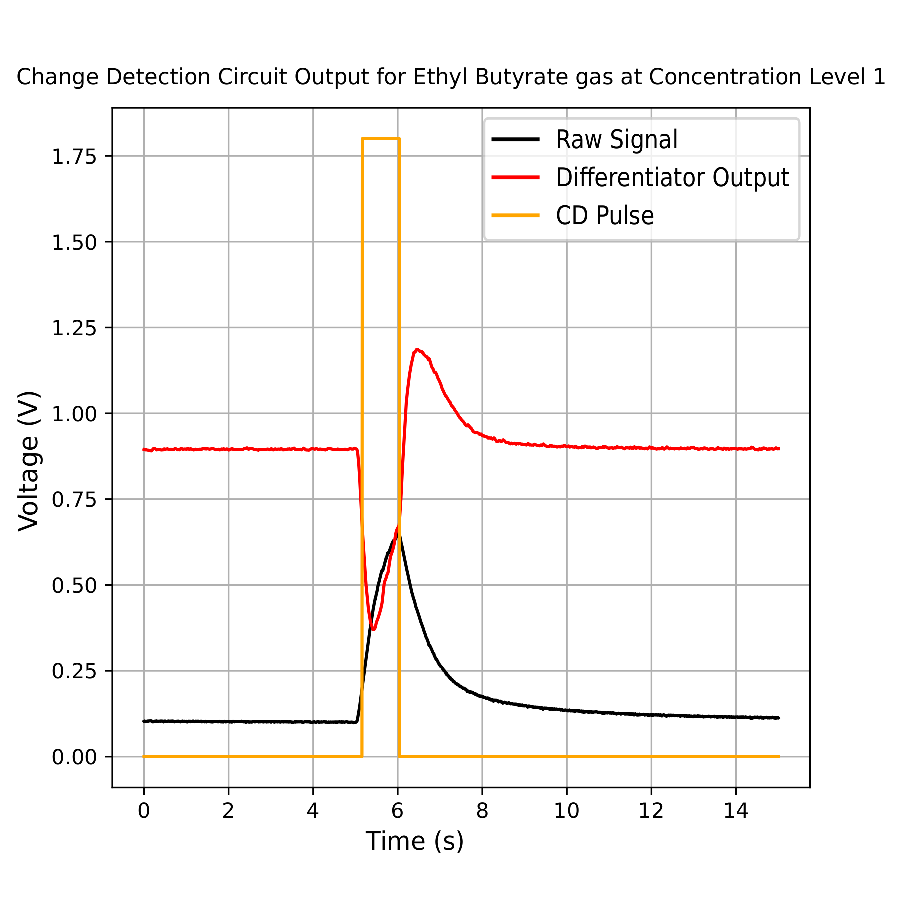}
\caption{Output of CD circuit for EB gas at concentration level 1}
\label{fig:cd_circuit}
\end{figure}

Figure 4 shows the output of the CD circuit for EB gas at concentration level 1. The red curve shows the inverting filtered difference circuit output. The time constants of the circuit are chosen in such a way that it has a high gain during the rising flank of the input signal and it reverts as soon as the input signal starts falling. Therefore, the CD pulse $Out_{CD}$ remains high during the rising flank and resets as soon as the input signal switches sign.

\subsubsection{The Exposure Measurement Circuit}

The exposure measurement stage, in which the input signal is integrated and compared with a threshold, is preceded by a high-pass filtering stage. The high-pass filter formed by resistor R3, capacitor C3, and the unity gain op-amp U3 removes low-frequency components from the MOx signal V1 which might encode some old gas information (especially for the MOx sensors with large time constants) and allows high-frequency transients to pass through it which usually encodes the current gas information. \par

The high-pass filtered MOx signal is provided to the exposure measurement stage through a pass-gate formed by PMOS transistor M1 and NMOS transistor M2. The gates of transistors M1 and M2 are controlled by the CD pulse $Out_{CD}$ and the inverted CD pulse $\overline{Out_{CD}}$ respectively. In the absence of a CD pulse, the output of the pass-gate is reset to an adjustable and arbitrary DC offset voltage set by potentiometer X3 through a reset transistor M3. Whenever the CD pulse is on, the reset transistor M3 is switched off and the pass-gate output is provided to the non-inverting integrator U4. In this way, the exposure measurement is triggered by the CD pulse. \par

To make sure that the exposure measurement system produces a sufficiently large response even for very weak MOx signals, the gain of the EM system has been chosen very high. The integrator output in the EM circuit saturates whenever it is operated in isolation and a signal with a non-zero mean is provided to its input terminal. In the complete front-end circuit shown in Figure 4, its operation is controlled by the CD pulse using pass-gate transistors M1 and M2, and the transistor switch M4 connected to the integrator U4. Therefore, the integrator output never saturates in this complete circuit. \par

The integrator U4 integrates the high-pass filtered sensor signal and the integrator output is compared with an adjustable threshold set by the potentiometer X4. Whenever the integrator output crosses this threshold, an exposure measurement pulse $Out_{EM}$ is generated. When the CD pulse $Out_{CD}$ goes off, the transistor M4 turns on which resets the integrator output. The EM pulse is reset whenever the integrator output goes below the threshold set by the potentiometer X4. \par

Figure 5 shows the output of the EM circuit for EB gas at concentration level 1. The blue curve is the output of the pass-gate which is a high-pass filtered MOx signal shifted by some DC offset. The green curve is the output of the integrator. The threshold value for the EM circuit is chosen in such a way that the EM pulse is always generated during the rising phase of the integrator output. In this way, the timing of the EM pulse will be different for a particular gas at different concentration levels. \par

\begin{figure}[h]
\includegraphics[width=\linewidth]{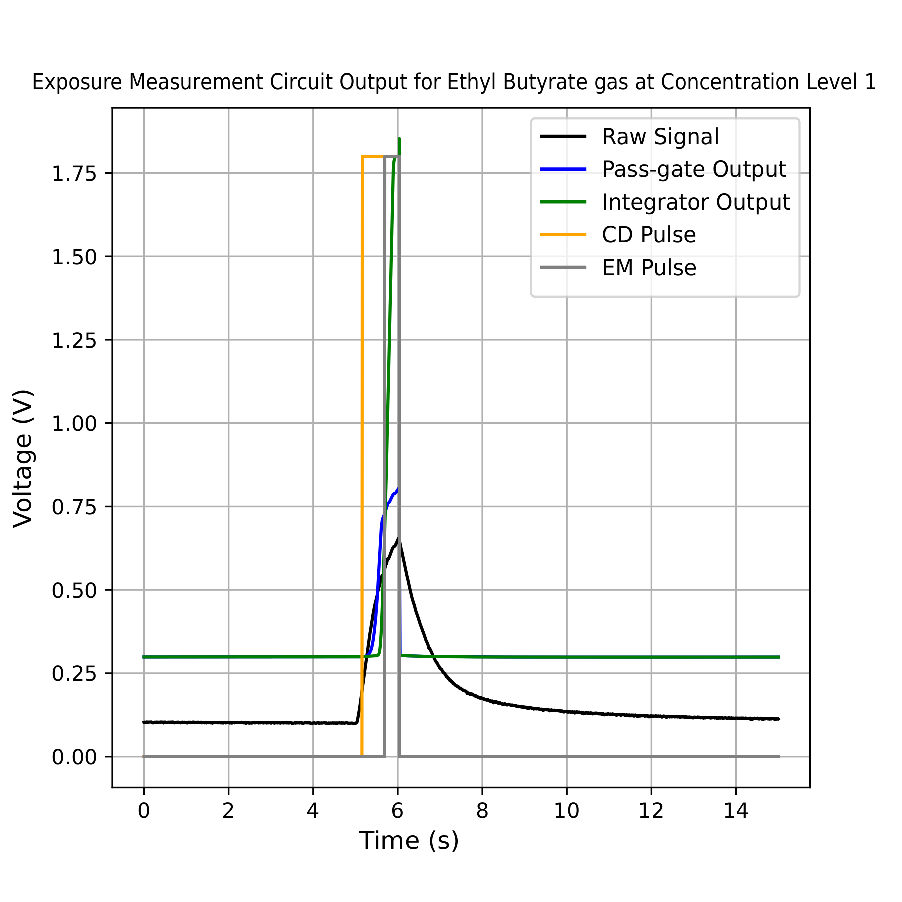}
\caption{Output of EM circuit for EB gas at concentration level 1}
\label{fig:em_circuit}
\end{figure}

\subsection{Sensor Array} 
For many applications such as air quality monitoring \cite{Srivastava2003}, for example, the exact identity of the gas whose concentration level we want to measure is not known in advance. In such applications, an array of MOx sensors with different sensitivities to different gases would provide much more detailed information and should allow us to decode the gas identity along with its concentration level through the combinatorial response of different sensors.  \par

\begin{figure}[h]
\includegraphics[width=\linewidth]{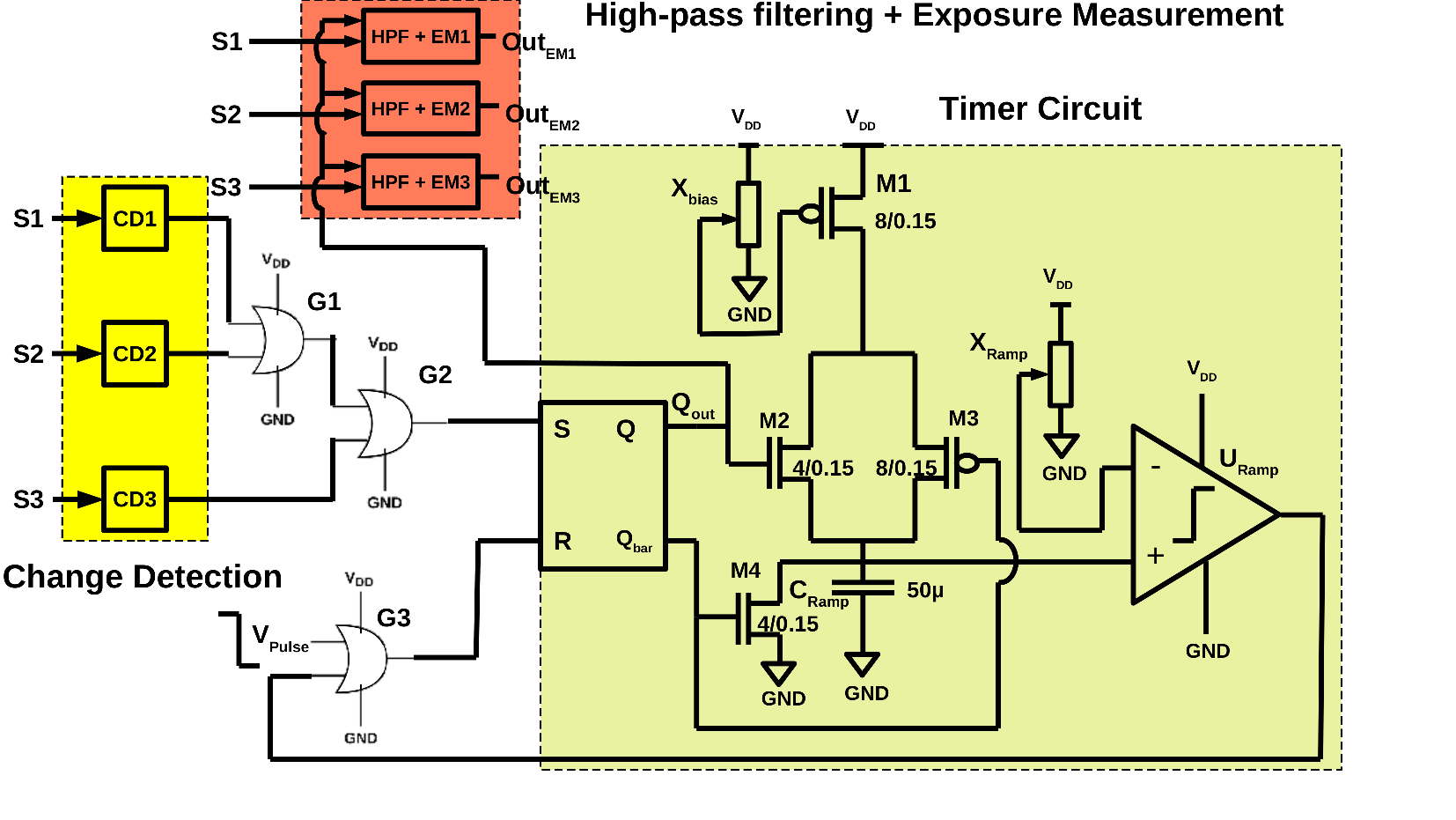}
\caption{The analog front-end for a 3-sensor array}
\label{fig:circuit_front_end_array}
\end{figure}

Figure 6 shows the front-end circuit for the array of 3 MOx sensors of MiCS-6814, whose recordings are shown in Figure 1. Similar to the case of a single gas sensor, this circuit also contains the change detection stage (CD1, CD2, and CD3), high-pass filtering (HPF), and exposure measurement stage (EM1, EM2, and EM3) for each sensor. The high-pass filtering stage is identical for each sensor in the array. S1, S2, and S3 are the voltage drops across load resistors connected in series with sensors 1,2, and 3 respectively which are fed to the change detection and the high-pass filtering stage of the corresponding sensor. Each change detection circuit generates the change detection pulse independently of each other, and the timing of each CD pulse depends on the signal generated by its respective gas sensor. In the case of a single gas sensor, the exposure measurement is triggered by the CD pulse. If we trigger the exposure measurement of each sensor in an array by its corresponding CD pulse, the exposure measurement for different sensors will start asynchronously, and the start of the exposure measurement for each sensor will depend on the timing of its corresponding CD pulse. This implies that there will be no exposure measurement for the sensor for which there is no CD pulse. Also, we need to transmit the timings of CD pulses of all sensors in the array along with their corresponding EM pulse timings to the inference stage. This transmission process has to be very clean so that the EM pulse timing of every sensor is transmitted along with its own CD pulse timing and not with the CD pulse timing of another sensor to make a proper inference which becomes a time-consuming process when we have a large number of sensors in the array because CD and EM pulses of different sensors are generated asynchronously. For a large sensor array, it will therefore be more bandwidth-efficient to start the exposure measurement of all sensors simultaneously by a global trigger pulse and then compute the timings of EM pulses of all sensors with respect to the timing of this global trigger pulse. \par

\subsubsection{The Timer Circuit}
The change detection pulses of 3 sensors are fed to OR gates G1 and G2 to obtain a resultant pulse $CD_{out}$. Because of the OR operation, this resultant pulse will be generated even if there is a CD pulse on any one of the sensors. This resultant pulse $CD_{out}$ can be used to trigger the exposure measurement in all sensors. But in case of any glitch in the OR gates or if we are in a scenario where there are rapid concentration changes, another exposure measurement might start in the middle of the previous exposure measurement which could lead to unreliable gas concentration measurement at any specific time. Therefore, it is important to store the $CD_{out}$ pulse and fix its duration. In this way, the duration of each exposure measurement will be fixed and it will not be affected by the glitch in OR gates or by any sudden concentration changes in that duration. It also provides a time-out function for all ongoing exposure measurements so that all exposure measurements are finished at the same time. \par

\begin{figure}[h]
\includegraphics[width=\linewidth]{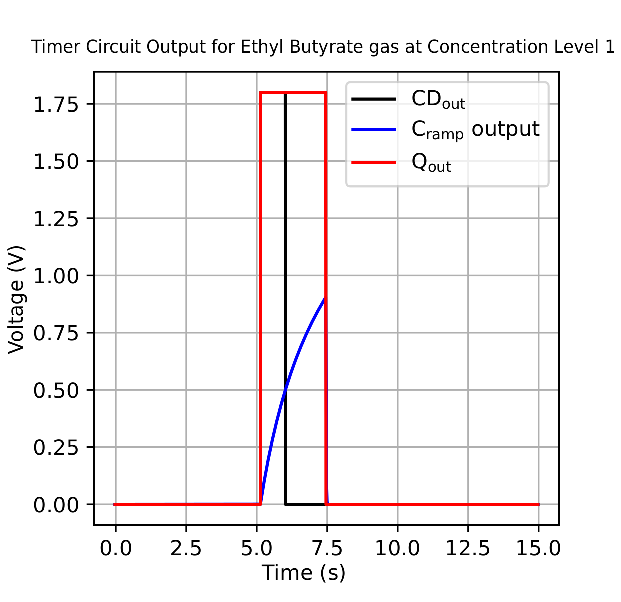}
\caption{Output of timer circuit for EB gas at concentration level 1 }
\label{fig:output_timer_circuit}
\end{figure}

The timer circuit shown in Figure 6 is used to fix the start time and the duration of exposure measurement for all sensors in the array so that the exposure measurement starts and ends synchronously for all sensors. The timer circuit contains an SR Latch to store the $CD_{out}$ pulse and a ramp generator circuit to adjust its width. When $CD_{out}$ is off in the beginning, the SR latch should be in the reset mode. This has been ensured by a pulse input $V_{pulse}$ and the OR gate G3. The pulse $V_{pulse}$ has been chosen in such a way that it is in the ON state for a few milliseconds in the beginning just to reset the latch.  It then goes to the OFF state and the SR latch remains latched to the reset state. Whenever there is a change detection pulse for any sensor in the array, the $CD_{out}$ pulse goes high and it sets the SR latch. Once the SR latch is set, it turns on the pass-gate formed by the NMOS transistor M2 and the PMOS transistor M3. When the pass-gate turns on, the PMOS transistor M1 starts charging the capacitor $C_{ramp}$ and it generates a ramp signal while charging (hence the name ramp generator). The bias voltage set by the potentiometer $X_{bias}$ for the gate of transistor M1 is set in such a way that M1 is not in a complete OFF state and the controlled current provided by M1 to $C_{ramp}$ is just enough to generate a ramp signal. Once this ramp signal reaches the threshold voltage set by the potentiometer $X_{Ramp}$, the comparator $U_{Ramp}$ generates a reset pulse which resets the SR latch through the feedback signal provided to the OR gate G3. When the latch resets, the transistor M4 turns on and discharges the capacitor $C_{ramp}$. The resetting of the SR latch ends the ongoing exposure measurement and the circuit is ready for another exposure measurement with the arrival of a new $CD_{out}$ pulse. \par

Figure 7 shows the output of the timer circuit for EB gas at concentration level 1. The black pulse is the $CD_{out}$ pulse which turns on the SR latch. Once the SR latch turns on, the latch output $Q_{out}$ goes high which charges the $C_{ramp}$ capacitor. The voltage across $C_{ramp}$ (shown by the blue curve) rises with capacitor charging. Once this voltage reaches the threshold level (which is 0.9 V here), the latch is reset by the comparator $U_{ramp}$, and the $Q_{out}$ pulse goes low. The start timing of the $Q_{out}$ pulse depends on the CD pulse generated by the fastest sensor. But its width is only decided by the charging time of the capacitor $C_{ramp}$ which is fixed in a particular setting and is independent of the CD pulses generated by individual sensors in the array. Therefore, the exposure measurement of all sensors triggered by the $Q_{out}$ pulse will start at the same time and its duration will be the same for all sensors irrespective of whether any sensor is fast or slow. As the start timing of the exposure measurement pulse will be different for different sensors depending on their sensitivities to a given gas, the gas concentration information from each sensor is encoded in the timing of the exposure measurement pulse relative to the timing of the global trigger pulse. \par

\section{Results}

\subsection{Single Sensor}

The proposed circuit for the single sensor was simulated and differences in timings of activation of CD and EM pulses were measured for all gases at all five different concentration levels. Figure 8 shows the plot of the inverse of the mean time difference over all trials for CD and EM pulse activation with respect to the concentration level of each gas. Error bars at each concentration level represent the standard deviation over all 20 trials. It can be observed that the time difference between the rising edges of CD and EM pulses varies inversely with gas concentration. In this way, the gas concentration information is encoded in the timings of analog spikes generated from two separate pathways, analogous to the spiking output of the mammalian olfactory bulb. To convert the finite duration analog pulses into short spikes as observed in biology, we can pass the CD and EM pulses through an edge detector circuit that generates fixed-duration spikes only at the positive edge of these pulses. These spikes can then be transmitted at the inference stage. \par

\begin{figure}[h]
\includegraphics[width=\linewidth]{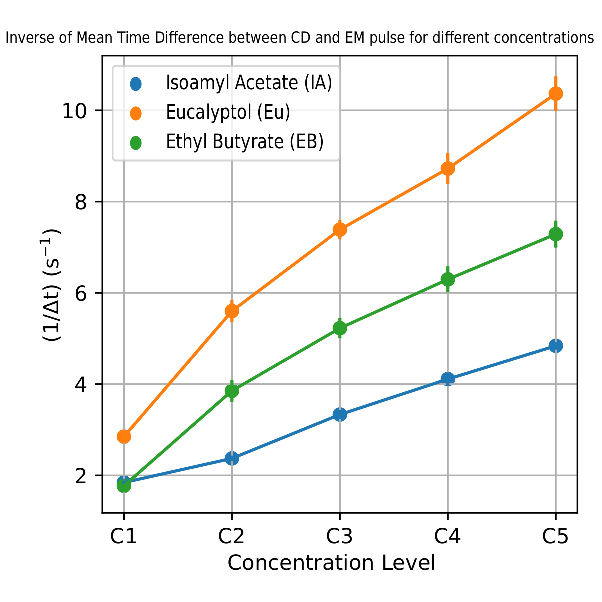}
\caption{Inverse of the time difference between CD and EM pulse activation with respect to concentration level. Dots and error bars represent the mean and standard deviation across 20 trials.}
\label{fig:inv_TD_conc_circuit_output}
\end{figure}

As a particular MOx sensor exhibits different sensitivity for different gases, the variation rate is not the same for all gases. There might not be a significant variation in the time difference between CD and EM pulses with respect to the concentration level for those gases for which the given MOx sensor has very low sensitivity. Therefore, the sensory front-end for a single sensor is only capable of measuring the concentration level of a known gas for which the MOx sensor has high sensitivity. \par

\subsection{Sensor Array}

The proposed sensor array circuit was simulated for all the MOx recordings shown in Figure 1 and the difference in timings of the global trigger $Q_{out}$ pulse and EM pulses ($\Delta{t}$) of all sensors were measured for all of the recordings. \par

\begin{figure}[h]
\includegraphics[width=\linewidth]{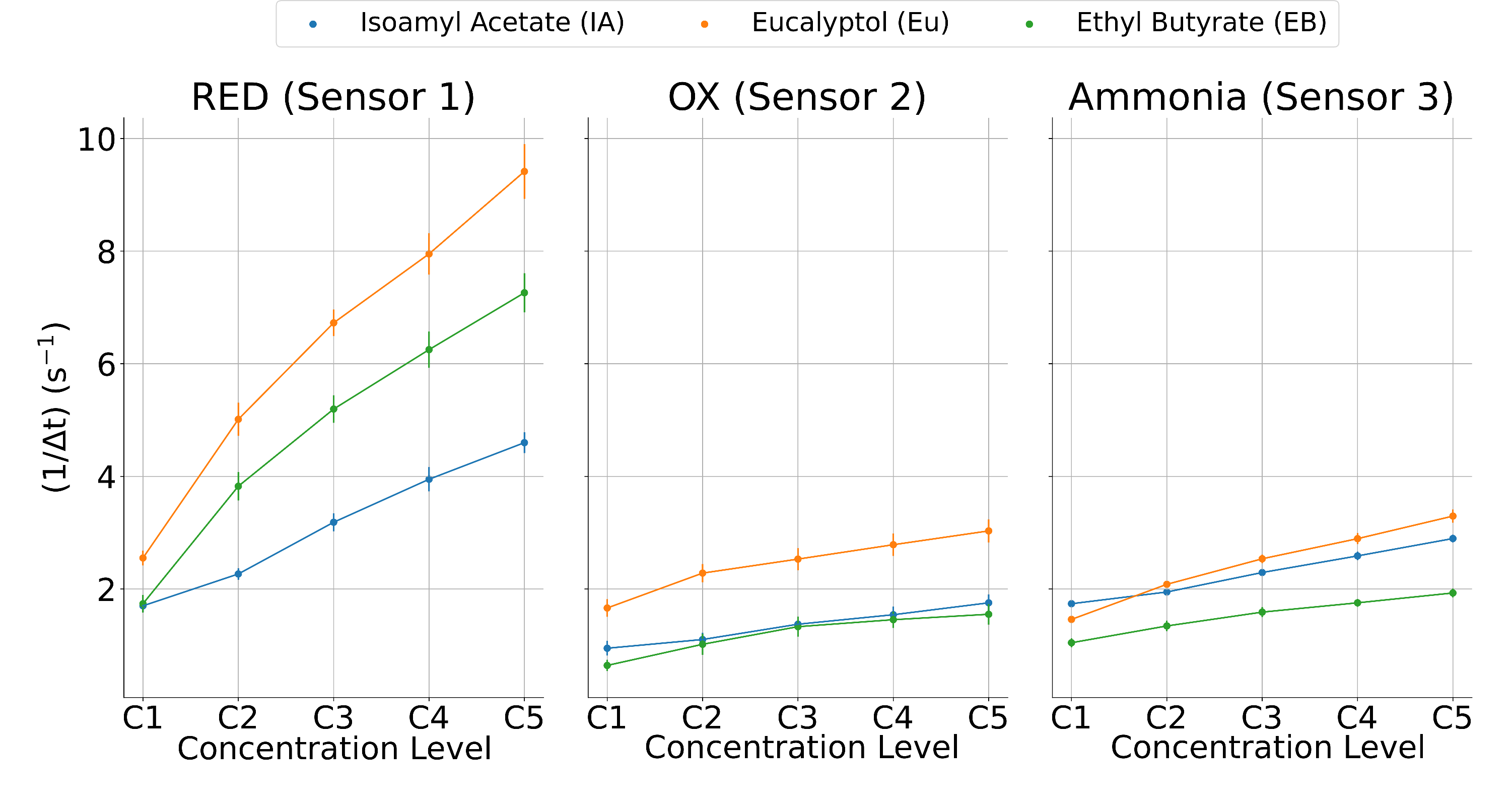}\caption{Inverse of the time difference between $Q_{out}$ and EM pulse activation with respect to concentration level for 3 sensors of MiCS-6814. Dots and error bars represent the mean and standard deviation across 20 trials respectively. For 8 trials of EB gas at C1, the EM pulse was not observed for sensor 2 and therefore, these trials for this sensor were discarded.}
\label{fig:inv_TD_conc_circuit_mox_array}
\end{figure}

Figure 9 shows the plot of the inverse of the mean time difference between $Q_{out}$ and EM pulse activation over all trials for 3 sensors of MiCS-6814. It can be seen that the time difference between $Q_{out}$ and EM pulse activation follows a similar trend with respect to the concentration level of gas pulses for all sensors. However, the variation of 1/$\Delta{t}$ with respect to the concentration level (slope) is very different for 3 sensors. The RED sensor is the fastest of all sensors and exhibits a small time constant for all tested gas pulses. Therefore, the variation rate is the highest for this sensor for all gases, as can be seen in Figure 9. The Ammonia sensor has the least sensitivity of all sensors for all the tested gases. Due to the larger time constant for all gases, the variation rate is the lowest for this sensor. It would be good to note that the order of time differences for multiple sensors in a sensor array is not always maintained. This implies that for a given sensor, the plot obtained in Figure 9 for one gas might intersect with the plot obtained for another gas for some concentration levels. This could be due to nonlinearities in the sensor sensitivity. The circuit readout for this 3-sensor array circuit provides us with a concentration vector of length 3 for a gas at a particular concentration. This concentration vector will be different for different gases and will also be different for the same gas at different concentration levels. Therefore, both the gas identity as well as its concentration level can be decoded at the inference stage through this obtained concentration vector. \par

Using MOx sensors of different sensitivities for a particular gas in the sensor array offers an advantage for this circuit over the single sensor circuit. For example, if a MOx sensor in the array has very low sensitivity for any gas at its lowest concentration level, it could be that there would be no EM pulse at the circuit output of this MOx sensor. If there are other MOx sensors in the array with comparatively higher sensitivities for this gas at its lowest concentration, the concentration vector obtained from the circuit readout of these sensors will help us to decode the concentration level and identity of the gas. In this way, other MOx sensors compensate for any low-sensitive MOx sensor in the array. \par

\section{Conclusion and Discussion}

This work presents simple analog circuit implementations for encoding gas concentrations in the timings of spikes obtained on two parallel pathways, inspired by the mechanism of a mammalian olfactory bulb. These circuits generate output only when there is some significant change in the environment due to gas injection, which is in contrast to ADCs that sample the gas sensor data even when there is no useful information. Therefore, the proposed circuits for a single sensor and a sensor array can be data-efficient substitutes for ADCs that can be interfaced with a MOx sensor and sensor array respectively. \par

Apart from the olfactory bulb, parallel processing streams also exist in the preprocessing stations of other sensory modalities such as the retina of the eye \cite{Callaway2005, Dacey1996, Reinhard2021, Kaplan1986}, cochlea of the ear \cite{Johnson2015, Chirila2007, Caird1983}, and also in the mammalian tongue \cite{Roper2009} and the skin \cite{Wang2021, Gardner2010, Hao2015}. These findings suggest that separating different stimulus features into separate channels at the receptor level before transmission into higher brain regions for processing seems to be a common and important property in all sensory modalities. Network simulations performed by Geramita et al. \cite{Geramita2016} prove that tufted cells perform best in odor discrimination at low concentration ranges and mitral cells perform best at high concentration ranges. The parallel combination of MCs and TCs discriminates odor best at intermediate concentration ranges and second best at low and high concentration ranges. This finding indicates that the parallel pathway system facilitates odor discrimination across a wider range of concentrations as compared to single pathway systems. \par

The circuits described in this work currently require manual adjustment of potentiometers for comparator thresholds, which could be cumbersome for a large sensor array. It would be of interest in the future to investigate some automated calibration strategies \cite{Leoni2018} \cite{Buhry2009} for adaptive comparator thresholds, which will reduce the manual operation and hence the operation of these circuits can be scaled for bigger e-nose systems. \par

The analog circuits for a single gas sensor and the gas sensor array have been tested for MOx recordings, which have been collected over 12 hours. Therefore, the trial-to-trial variability observed in these recordings due to the interaction of chemical analytes at the sensing film, because of long-term usage of sensors (first-order sensor drift \cite{Vergara2012}) is small. Also, these recordings have been obtained in a controlled environment setting with a regular airflow and therefore have not been impacted by the changes in environmental conditions, or noise in the measurement delivery system (second-order drift \cite{Vergara2012}). \par

Long-term exposure to gases often makes MOx sensors prone to first-order drift, which shifts the baseline resistance of MOx sensors \cite{dennler2022drift}. Considering long-term experiments, we expect that our proposed circuits are resilient to this first-order baseline drift because the circuits use the band-pass filter as one of the subcircuits to cancel out this baseline shift. Additional long-term experiments are needed in the future to fully assess how drift affects these circuits. Our conviction is that the MOx array circuit is much more robust in terms of drift effects than the single sensor circuit. This is because the circuit readouts from other cleaner sensors in the array would compensate for the drift-ridden sensor. \par 

The proposed sensor array front-end has been designed and tested only for a 3-sensor array. To further validate the robustness of this circuit to drift effects and to deploy it in complex environments comprising other interfering gases apart from the target gas, the circuit should be tested for a large sensor array comprising MOx sensors of a diverse range of sensitivities to a particular gas. However, scaling up the circuit for large arrays would create additional issues such as less area efficiency due to timer circuit and OR gates, and false positives due to cross-sensitivity of gas sensors \cite{Mei2024}. Reducing the circuit complexity for large array systems to create area-efficient e-nose front-ends and further circuit modifications for reducing the number of false positives are valuable future directions for this work. \par

The proposed circuits are environment-sensitive and can only be used in a scenario where sensors are exposed to a single gas pulse of varying concentrations in constant airflow.  However, a more realistic environment contains gas plumes that are formed as molecules of a single gas or a mixture of gases are dispersed by the wind away from the source \cite{Murlis1992}. In such an environment, the gas concentration is not fixed but changes dynamically in both spatial and temporal domains because of turbulence \cite{Schmuker2016}. Therefore, the response of a MOx sensor in these environments could be very different from that in a controlled single pulse environment \cite{Schmuker2016, Burgues2019, Nakamoto2008}. It would be of interest in the future to analyze the MOx sensor data recorded for more natural plumes of a single gas or gas mixtures with time-varying concentrations and to do further design modifications in the proposed circuits \cite{Rastogi2024temporalodor} if needed to make them work properly in more realistic environments. \par

\bibliographystyle{ieeetr}

\bibliography{ref}

\vspace{12pt}

\section*{Author contributions statement}
\textbf{Shavika Rastogi:} Conceptualization, Methodology, Formal analysis, Software, Investigation, Writing - Original draft preparation. \textbf{Nik Dennler:} Data curation, Writing - Reviewing and Editing. \textbf{Michael Schmuker:} Supervision, Funding acquisition, Writing - Reviewing and Editing. \textbf{André van Schaik:} Conceptualization, Supervision, Funding acquisition, Writing - Reviewing and Editing.

\end{document}